\documentclass[runningheads]{llncs}
\usepackage{graphicx}
\usepackage{hyperref}
\usepackage{mathtools}
\usepackage{amsmath}
\begin{document}
\title{Introducing the Attribution Stability Indicator:\\a Measure for Time Series XAI Attributions}
\titlerunning{Introducing the Attribution Stability Indicator}
%
\author{Udo Schlegel\inst{1}\orcidID{0000-0002-8266-0162} \and
Daniel A. Keim\inst{1}\orcidID{0000-0001-7966-9740}}
\authorrunning{Schlegel \& Keim}
%
\institute{University of Konstanz, Universitätsstraße 10, 78464 Konstanz, Germany
\email{\{u.schlegel,daniel.keim\}@uni-konstanz.de}}
\maketitle              
\begin{abstract}
Given the increasing amount and general complexity of time series data in domains such as finance, weather forecasting, and healthcare, there is a growing need for state-of-the-art performance models that can provide interpretable insights into underlying patterns and relationships.
Attribution techniques enable the extraction of explanations from time series models to gain insights but are hard to evaluate for their robustness and trustworthiness.
We propose the Attribution Stability Indicator ($ASI$), a measure to incorporate robustness and trustworthiness as properties of attribution techniques for time series into account.
We extend a perturbation analysis with correlations of the original time series to the perturbed instance and the attributions to include wanted properties in the measure.
We demonstrate the wanted properties based on an analysis of the attributions in a dimension-reduced space and the $ASI$ scores distribution over three whole time series classification datasets.

\end{abstract}
\section{Introduction}

Artificial intelligence (AI) has become an indispensable part of our everyday lives. 
We encounter it in various forms, from the tailored advertisements we receive on social media to the conversational AI (chatbots) that utilize deep neural networks to answer user and customer queries. 
However, as deep neural network models grow in complexity, understanding the rationale behind their decisions becomes increasingly challenging~\cite{guidotti_survey_2018}. 
This lack of interpretability can have grave repercussions in critical domains like finance, healthcare, and transportation, potentially resulting in financial losses, medical errors, or even loss of life due to incorrect decisions made by intricate models~\cite{rudin_stop_2019}. 
To address these concerns, a promising solution lies in the adoption of explainable artificial intelligence (XAI). 
XAI aims to shed light on the inner workings of complex models and the factors influencing their decision-making~\cite{guidotti_survey_2018}. 
Within this field, one area of particular interest is time series data, characterized by its sequential nature and interdependencies between observations. 
As the volume of data generated by sensors increases and complex models are applied to tackle more tasks, the need for explainability in time series analysis becomes evident~\cite{theissler_explainable_2022}.

In recent years, there has been a growing focus on developing explainable artificial intelligence (XAI) techniques specifically designed for time series data~\cite{theissler_explainable_2022}.
These techniques often rely on the concept of attributions, which aim to uncover the contributions of individual features and time points to the overall predictions made by a model~\cite{theissler_explainable_2022}. 
By revealing which parts of the input data are most relevant to the output, attributions assist users in understanding the decision-making process of the model~\cite{schlegel_time_2021}. 
However, evaluating these attributions poses a non-trivial challenge~\cite{schlegel_empirical_2020}. 
To tackle this challenge, perturbation analysis has emerged as a promising evaluation technique for assessing the quality of explanations in time series data~\cite{schlegel_towards_2019,simic_perturbation_2022}. 
Perturbation analysis involves systematically modifying the input data and assessing the resulting impact on the attributions generated by XAI methods~\cite{schlegel_empirical_2020}. 
By perturbing the input data, it becomes possible to evaluate the robustness of the explanations provided by XAI methods~\cite{theissler_explainable_2022}. 
For instance, Schlegel et al.~\cite{schlegel_empirical_2020} begin to explore the evaluation of attributions in detail and reveal a gap for an attribution measure incorporating the distance of the original and perturbed samples.

We propose the Attribution Stability Indicator ($ASI$), which incorporates various similarity measures between time series, attributions, and predictions.
We incorporate a requirements analysis to reveal the properties of an attribution technique measure to collect important aspects for working approaches in the direction of measuring.
Based on this requirement analysis, we propose five essential factors: a class flip, prediction probability changes, attribution distances, time series perturbation distances, and user-based weighting.
$ASI$ then incorporates our identified requirements by using the class flip as a binary indicator, the Jensen-Shannon distance~\cite{endres_new_2003} for the prediction probabilities, the Pearson correlation~\cite{galton_regression_1886} for the change in the attributions and for the change in the time series as measures.
As the ranges from these measures are between zero and one, we further introduce weighting to steer the measure in a user-wanted direction.
We also provide default values for these hyperparameters.
While the overall $ASI$ approach is quite general, we focus on time series to demonstrate applicability and incorporate existing working methods from the literature~\cite{aghabozorgi_time_2015}.

Thus, we contribute (1) a requirements analysis of the properties a measure for attribution techniques needs to evaluate these, (2) a perturbation analysis-based measure to include time series and attribution properties into the evaluation process, and (3) a small-scale evaluation to compare various attribution techniques.
At first, we relate our measure to other approaches to XAI evaluation.
Next, we break down the evaluation of attributions for time series using a perturbation analysis and identify essential requirements for attributions we want to measure.
Afterward, we introduce the attribution stability indicator parts and definitions based on these requirements.
Then, we use the \textsc{FordA}, \textsc{FordB}, and \textsc{ElectricDevices} datasets~\cite{dau_ucr_2019} on time series classification to demonstrate the $ASI$ measure against a flip like, e.g., Schlegel et al.~\cite{schlegel_towards_2019} proposed.
Lastly, we conclude our proposed approach and motivate further extensions.

Results and source code of the experiments is online available at:\\
{\small\href{https://github.com/visual-xai-for-time-series/attribution-stability-indicator}{https://github.com/visual-xai-for-time-series/attribution-stability-indicator}}

\section{Related Work}

Explainable AI (XAI) has made significant progress in recent years, driven by various surveys~\cite{guidotti_survey_2018,adadi_peeking_2018} and techniques such as LIME~\cite{ribeiro_why_2016} and SHAP~\cite{lundberg_unified_2017}. 
XAI is growing in many directions, starting from computer vision~\cite{zeiler_visualizing_2014} and moving towards time series~\cite{theissler_explainable_2022}.
However, evaluating explanations remains a nascent area, with limited efforts to benchmark different techniques against one another~\cite{hooker_benchmark_2019}. 
Several studies have begun to compile a range of evaluation techniques~\cite{mohseni_multidisciplinary_2021}, which have been categorized into five dimensions: mental model, explanation usefulness and satisfaction, user trust and reliance, human-AI task performance, and computational measures. The initial dimensions primarily revolve around evaluating explanations in collaboration with humans, making them significantly influenced by human factors. 
On the other hand, the computational measures focus solely on the automatic evaluation of explanations, excluding human factors~\cite{mohseni_multidisciplinary_2021}. 
Here, we examine the computational measures, explicitly exploring the fidelity of the attribution technique applied to the model toward an analysis of the fitness of the explanation. 

Mainly for time series, we have various levels on which time series model decisions can be explained~\cite{theissler_explainable_2022}.
We focus on time-point explanations to work directly on the raw input time series and attributions based on the taxonomy from Theissler et al.~\cite{theissler_explainable_2022}.
Previous research conducted by Schlegel et al.\cite{schlegel_towards_2019}, as well as others\cite{simic_perturbation_2022,mercier_time_2022,theissler_explainable_2022}, has demonstrated the effectiveness of attribution techniques such as LIME~\cite{ribeiro_why_2016}, SHAP~\cite{lundberg_unified_2017}, LRP~\cite{bach_pixel_2015}, GradCAM~\cite{selvaraju_grad_2017}, Integrated Gradients~\cite{sundararajan_axiomatic_2017}, and others~\cite{schlegel_ts_2021} in generating meaningful attributions from time series models.
However, in most cases, these attributions are evaluated using computational measures without further scrutiny, as observed by Mercier et al.~\cite{mercier_time_2022}. 
This limitation calls for a more in-depth examination computationally and by humans to gain deeper insights into the attributions.

Schlegel et al.\cite{schlegel_towards_2019} initiated their research by applying perturbation analysis to attribution techniques in TSC, demonstrating that techniques designed for images and text are also effective for time series data. 
Building on these initial experiments, they further refined their approach by incorporating additional perturbation functions, resulting in a more comprehensive evaluation of fidelity\cite{schlegel_empirical_2020}.
Mercier et al.\cite{mercier_time_2022} expanded on these perturbations by incorporating additional measures from the image domain, such as (in)fidelity and sensitivity\cite{yeh_in_2019}. 
Simic et al.\cite{simic_perturbation_2022} extended the methods proposed by Schlegel et al.\cite{schlegel_empirical_2020} by incorporating out-of-distribution values for the perturbation and provided guidelines for selecting attribution techniques and determining the appropriate window size for perturbation windows.
Similarly, Turbe et al.~\cite{turbe_interprettime_2022} further enhanced previous approaches by introducing an additional metric that improves the comparison of attribution techniques and their ability to demonstrate fidelity towards the underlying model.
However, none of these methods incorporate the distance in the attribution space and assume that these should also differ.

\section{Evaluation of Attributions for Time Series}

As a starting point for evaluating attributions for time series, we begin with the perturbation analysis described by Schlegel et al.~\cite{schlegel_empirical_2020,schlegel_deep_2023} and use the definitions and notions of Theissler et al.~\cite{theissler_explainable_2022} in their survey.

\subsection{Definitions for a Perturbation Analysis}

We define a time series classification dataset $D = (X, Y)$, where $X$ represents the time series samples and $Y$ denotes the corresponding time series labels. 
The set $X = \{ts_1, ts_2,\dots, ts_n\}$ consists of $n$ time series samples, each $ts_i$ containing $m$ time points represented as an ordered set of $m$ real-valued time points $ts = \{tp_1, tp_2,\dots, tp_m\}$. 
Here, $tp_1$ signifies the value at the $i$th time point of $ts$. 
The set $Y = \{l_1, l_2,\dots, l_n\}$ includes $n$ labels, with one label assigned to each time series sample.
Let $M(ts, \theta) = y'$ be a time series classification model that predicts the label $y'$ based on the input time series $ts$, using the parameters $\theta$ and for all time series in a dataset: $M(X, \theta) = Y'$. 
Additionally, we introduce $A(X, M, \theta)$ as an XAI technique designed to generate attributions for the time series data.
The attributions for $ts$ produced by $A$ are represented as $A(ts, M, \theta) = att$ and $att = \{a_1, a_2,\dots, a_m\}$. 
Here, $a_i$ corresponds to the attribution score assigned to the $i$th time point of $ts$. 
For all time series in a dataset: $A(X, M, \theta) = Att$ with $Att = \{att_1, att_2,\dots, att_n\}$. 
These attributions are generated using the classification model $M$ and the parameters $\theta$ associated with the technique.

We introduce a controlled perturbation function denoted as $g$, which operates on and modifies the dataset $X$ to facilitate perturbation analysis. 
Specifically, we define a perturbed time series dataset $X'$ as follows: $X' = g(X, Att, \xi)$.
The perturbation function $g$ alters the dataset $X$ based on the attributions $att$ and a threshold value $\xi$. The modification depends on the chosen function $g$, such as perturbing values to zeros.
The threshold $\xi$ can be manually specified or determined using another function, for instance, by setting it to the 90th percentile of the attributions. 
This way, attributions (e.g., $a_i$, representing the $i$th element) exceeding the threshold will be modified to a predetermined value, such as zero.

\begin{figure}[h!tb]
    \centering
    \includegraphics[width=\textwidth,trim={0 177mm 225mm 0},clip]{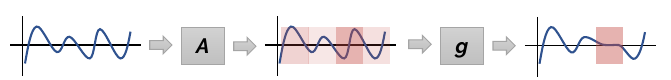}
    \caption{Starting from a time series $ts$, we use a selected attribution technique $A$ to get attributions. Based on the attributions, we use a selected perturbation function $g$ to set highly relevant time points, e.g., to zero. Further information in Schlegel et al.~\cite{schlegel_empirical_2020}.}
    \label{fig:pertrubation_analysis}
\end{figure}

In~\autoref{fig:pertrubation_analysis}, we illustrate the approach using zero perturbations on attributions with high values, showcasing the practical implementation of the technique.
To get more information on the perturbation strategies and the influences of using different perturbation strategies on various datasets, we suggest the work by Schlegel and Keim~\cite{schlegel_deep_2023}.

We evaluate the performance of the model $M$ on both the original dataset $X$ and the perturbed dataset $X'$ by obtaining the predictions $M(X) = Y'$ and $M(X') = Y''$, respectively.
In line with the findings of Schlegel et al. ~\cite{schlegel_empirical_2020}, we incorporate a quality metric denoted as $qm$, e.g., accuracy or a similar measure. 
This metric enables us to compare the performance of the model $M$ using the original dataset $X$ and the perturbed dataset $X'$.
In the context of time series classification, we assume that the quality metric $qm$ decreases as the original data changes, implying that the previously assigned labels are no longer accurate~\cite{schlegel_towards_2019}. 
Moreover, we also assume that an effective attribution technique, when applied to perturb the most relevant parts of the input data~\cite{hooker_benchmark_2019}, leads to a more significant decrease in performance.

\begin{figure}[h!tb]
    \centering
    \includegraphics[width=0.95\textwidth,trim={0 32mm 24mm 0},clip]{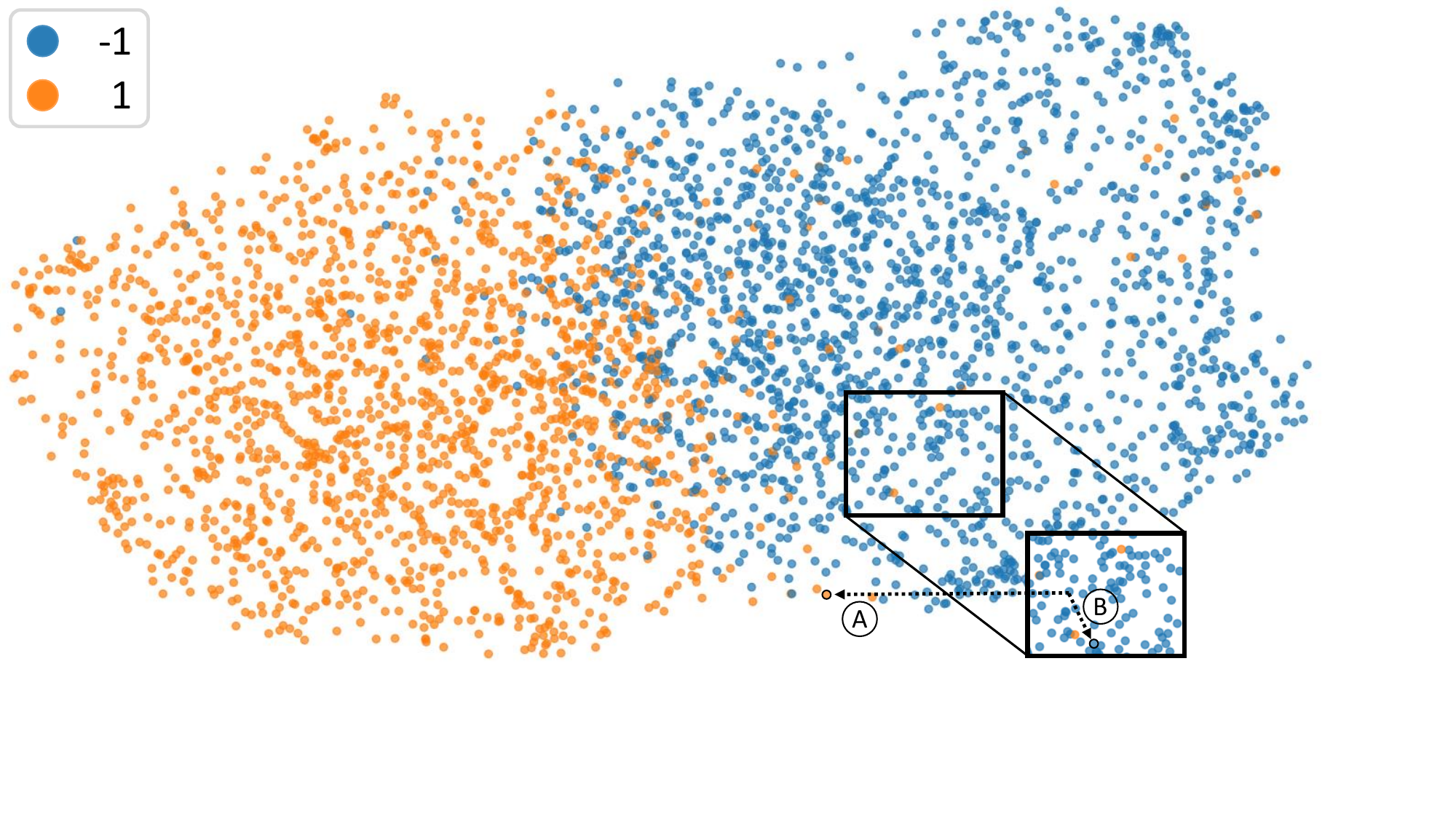}
    \caption{Projection of the \textsc{FordA} training data attributions of DeepLIFTSHAP on a CNN using UMAP. (A) is wanted after a perturbation, while (B) is not.}
    \label{fig:attribution-projection}
\end{figure}

Hence, Schlegel et al.~\cite{schlegel_empirical_2020} posit the following relationship:
\begin{equation}
qm(Y', Y) \leq qm(Y'', Y)
\end{equation}
Here, $qm(Y', Y)$ represents the quality metric for the predictions $Y'$ compared to the actual labels $Y$, while $qm(Y'', Y)$ corresponds to the quality for the predictions $Y''$ obtained from the perturbed dataset $X'$.
Depending on the $qm$, the performance drastically changes.
Thus, a more independent measure is needed.

\subsection {Requirements for a Measure on the Perturbation Analysis}

Previous works, such as Schlegel et al.~\cite{schlegel_empirical_2020}, demonstrate that a class change is one of the most wanted properties to determine an attribution technique's fidelity.
Thus, a class change in prediction is essential for working attributions as these reveal the highly relevant parts of the time series for the models' prediction (R1).
However, a binary decision on the class flip is sometimes not enough as a measure, as attribution techniques can present part-time working attributions that do not necessarily lead to a class flip.
Thus, we need to dig into the probabilities of the class prediction of the model and compare the original to the perturbed instance to find more minor differences in the perturbation changes (R2).

For a given dataset, we can calculate the attributions for every sample and use a projection technique to visualize the attributions of the whole dataset.
\autoref{fig:attribution-projection} presents the attributions extracted with Integrated Gradients for the training data of the \textsc{FordA} dataset applied to a CNN using the projection technique UMAP~\cite{mcinnes_umap_2018} (further information on the dataset and the model in~\autoref{sec:experiments}).
We have a two-class problem and want suitable attributions so that these attributions differ quite heavily for each class and build two clusters to generate a visual representation of the model.
\autoref{fig:attribution-projection} demonstrates such a scenario.
A working perturbation for a suitable attribution then changes the attribution of the perturbed instance toward the other cluster, as in (A).
Bad attribution techniques create perturbed instances with similar attributions, such as (B).
Thus, a promising measure must include the original and the perturbed instance attribution similarity (R3).

\begin{figure}[h!tb]
    \centering
    \includegraphics[width=0.9\textwidth,trim={0 125mm 40mm 0},clip]{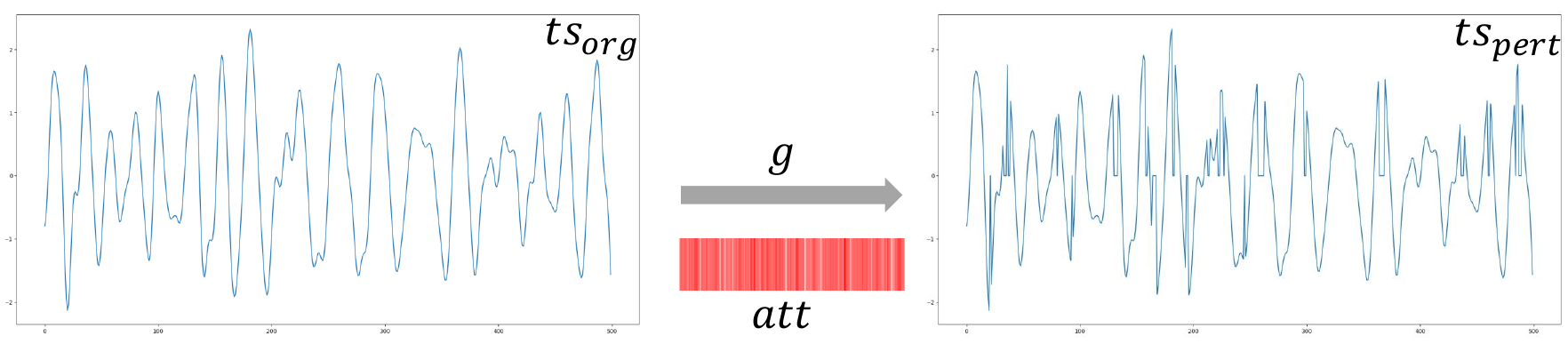}
    \caption{\textsc{FordA} time series with a change of a perturbation function $g$ with attributions $att$ form $ts_{old}$ to $ts_{new}$ using zero as perturbation value.}
    \label{fig:time-series-change}
\end{figure}
Contrary to the attributions, we want our perturbed instance to have a small distance to the original time series to change as little as possible and still get our prediction change.
As seen in~\autoref{fig:time-series-change}, the more time points we change using a perturbation function, the more jumps and cuts our novel time series has.
Depending on our black box, such a change can already lead to a change in the prediction.
However, we do not want to have many cuts and changes, so we want our measure to include the distance from the original to the perturbed instance to include a minimal change in our evaluation (R4).

Lastly, as we know that not every dataset is the same and some tasks have different requirements, we need to be able to weigh the factors we identified in our analysis.
E.g., for some applications, the change in prediction is much more critical to understanding the model's weaknesses. 
Thus, we want to have an attribution technique that reveals such flips.
In further scenarios, we want to investigate our attributions as visualizations with, e.g., a projection such as in~\autoref{fig:attribution-projection}, so we need to weigh our attributions heavier.
Thus, even with the previous requirements, we need to be able to include user necessities (R5).

\textbf{Summary --}
As we described before, a measure for the analysis of attribution techniques either on a single sample or on a whole dataset needs the following requirements: 
(1) change the predicted label (a class flip after the perturbation analysis based on the attribution), 
(2) a drastic change in the probabilities of the output of the model, 
(3) a significant distance between the original and the perturbed samples attribution due to a change in the prediction, 
(4) only minor changes between the original time series and the perturbed instance are needed to decision boundaries, and
(5) lastly, weighting helps to let users steer the different requirements based on their properties and needs.

\subsection{The Attribution Stability Indicator as a Measure}

On the perturbation, we propose the Attribution Stability Indicator ($ASI$) as another measure to not only use one single quality metric as, e.g., Schlegel et al.~\cite{schlegel_towards_2019} or Simic et al.\cite{simic_perturbation_2022} to include the requirements we identified.
To structure and introduce the parts of $ASI$, we define our approach to tackle the requirements and give some insights into the selected techniques.

\textbf{Requirement 1 --}
We start with the most straightforward requirement we have identified, targeting the classification change.
To get such a change, we use the selected label by the model with $x = ts$ to $M(ts) = y'_{pre}$ and the prediction after the perturbation $M(g(ts, att) = y'_{new}$.
Further, we use the function $b_f$ as we want to have a low score for a working technique:
\begin{equation}
    b_f(X, Y)= 
\begin{cases}
    1, & \text{if } \text{argmax}(X) = \text{argmax}(Y)\\
    0, & \text{otherwise}
\end{cases}
\end{equation}
With $X$ and $Y$ being $y'_{pre}$ and $y'_{new}$ respectively and argmax extracting the index of the largest value.
Leading us in the end to:
\begin{equation}
b_f(M(ts), M(g(ts, att))
\label{eq:bf}
\end{equation}

\textbf{Requirement 2 --}
Next, we target the change of the prediction probability distribution with the Jensen-Shannon distance~\cite{endres_new_2003}.
The Jensen-Shannon divergence (JSD) can be defined as:
\begin{equation}
JSD(P \parallel Q) = \frac{1}{2} \left( D_{KL}\left(P \parallel \frac{P+Q}{2}\right) + D_{KL}\left(Q \parallel \frac{P+Q}{2}\right) \right)
\label{eq:jsd}
\end{equation}
With $P$ and $Q$ as the two probability distributions and $D_{KL}$ as the Kullback–Leibler divergence.
By taking the square root, the JSD,~\autoref{eq:jsd}, can be used as a distance (Endres and Schindelin~\cite{endres_new_2003}) reforming the JSD to:
\begin{equation}
JS\_{dist}(P, Q) = \sqrt{\frac{1}{2} \left( D_{KL}\left(P \parallel m\right) + D_{KL}\left(Q \parallel m\right) \right)}
\label{eq:jsdd}
\end{equation}
With $m$ as the pointwise mean of $P$ and $Q$.
We use the $JS_{dist}$,~\autoref{eq:jsdd}, for the similarity between the old $p$ predicted probabilities of the model and the new ones $p_{pert}$.
As the $JS_{dist}$ already uses our wanted range from zero to one, we can use that value as it is.
However, as a one in the distance describes a working change in the distribution, we have to change the one to a zero in the working case, leading us to:
\begin{equation}
1 - JS_{dist}(M(ts), M(g(ts, att)))
\label{eq:1jsdd}
\end{equation}

We decided to use the Jensen-Shannon distance as we want to have a symmetric measure to combine it with other symmetric measures in our approach, such as the correlation for the time series and attributions.
However, other symmetric distribution measures, such as the Hellinger distance~\cite{hellinger_neue_1909} ($H_{dist}$) or the Bhattacharyya distance~\cite{bhattacharyya_measure_1943} (does not obey the triangle inequality), are also possible.
In a preliminary experiment between $JS_{dist}$ and $H_{dist}$, we observed that the $JS_{dist}$ emphasizes differences in the tails of the distributions, while the $H_{dist}$ is more sensitive to the central parts of the distribution.
So, the $JS_{dist}$ focuses more on outliers while $H_{dist}$ focuses more on the shape.
For our measure, the $JS_{dist}$ is more advantageous as we are more interested in small changes and not the overall dissimilarity of the shapes.

\textbf{Requirement 3+4 --}
For the attribution and time series similarity, we decided on the Pearson correlation coefficient from the original $ts$ to the perturbed instance $ts_{pert}$ as the values range from minus one to plus one.
We use the base definition for the Pearson correlation coefficient:
\begin{equation}
\rho(X,Y) = \frac{{\text{cov}(X, Y)}}{{\sigma_X \sigma_Y}}
\label{eq:rho}
\end{equation}
With $X$ and $Y$ being our $ts$ and $ts_{pert}$ respectively and $\text{cov}(X, Y)$ the covariance between them as: $\text{{cov}}(X, Y) = \frac{{\sum_{i=1}^{n} (X_i - \sigma_X)(Y_i - \sigma_Y)}}{{n-1}}$.
$\sigma_X$ and $\sigma_Y$ describe the means of $X$ and $Y$.
As our $\rho$,~\autoref{eq:rho}, is always between minus one and plus one, we use these borders to normalize the result to zero to one range to include it in the formula as it is.
Leading us to:
\begin{equation}
pcc(X,Y) = \frac{\rho(X,Y) + 1}{2}
\label{eq:pcc}
\end{equation}
And respectively for the attributions to:
\begin{equation}
pcc(att, A(g(ts, att), M))
\label{eq:pcca}
\end{equation}
And the time series with a change to highlight large distances:
\begin{equation}
1 - pcc(ts, g(ts, att))
\label{eq:pccts}
\end{equation}

We used the Pearson correlation coefficient due to its simplicity and interpretability. 
The Pearson correlation is a straightforward and easy-to-understand measure.
The calculation of the Pearson correlation involves simple mathematical operations, which makes it computationally efficient.
Also, the Pearson correlation provides a clear interpretation of the strength and direction of the relationship between two time series. 
The correlation coefficient allows us to quickly compare different pairs of variables and assess their level of association. 
However, other correlation measures are also possible with a bound between zero and one or some fixed bound and normalization.
Generally, any distance measure with bounds, which can be modified to zero and one to enable uniform weighting and direction between the different internal measures, can be incorporated into the measure.
We still advise using a correlation to include 

\textbf{Requirement 5 --}
To enable users to gain influence into the steering of the measure, we introduce a weighting term $W$ with a weighting between zero and one $w_i\in \{0,1\}$ to get to our weighting factor for our four other requirements $W = (w_1, w_2, w_3, w_4)$.
E.g. if a user wants to find attribution techniques that can be interpreted using projections, the weight for the attribution distance can be adjusted to a higher weight.
Thus, the case of (A) in~\autoref{fig:attribution-projection} is much more likely than case (B) to gain low scores in a measure using the previous requirements.
Also, through such a weighting, we can mimic the raw flip amount using $W_{\text{raw}} = (1, 0, 0, 0)$ for our approach.

\textbf{Attribution Stability Indicator ($ASI$) --}
To put everything together with our weighting $W = (w_1, w_2, w_3, w_4)$, we get to:
\begin{equation}
\begin{split}
ASI(M, ts, att, W) &= \frac{1}{4} (w_1 \times b_f(M(ts), M(g(ts, att)))\\
&+ w_2 \times (1 - JS_{dist}(M(ts), M(g(ts, att))))\\
&+ w_3 \times pcc(att, A(g(ts, att), M))\\
&+ w_4 \times (1 - pcc(ts, g(ts, att))))
\end{split}
\end{equation}
First, we include our binary flip, next our probability distribution similarity, then our correlation between attributions, and lastly, our time series correlation.
The weighting enables the focus on specific aspects of the requirements heavier and users to steer the whole score in a desired direction.
The function ranges from zero to one, with a lower score presenting better results.
We also discussed incorporating the Euclidean distance or dynamic time warping for the time series and similar distances.
However, these can have arbitrary values; in some cases, a comparison is rather not understandable.

\section{Experiment Setup and Results}
\label{sec:experiments}

Our experiment analyzes data characteristics using three of the most exhaustive datasets (\textsc{FordA}, \textsc{FordB}, \textsc{ElectricDevices}) from the UCR benchmark dataset collection~\cite{dau_ucr_2019}.
While we specifically examine these datasets, it is essential to note that our approach applies to any time series classification dataset.
The \textsc{FordA} and \textsc{FordB} datasets consist of sensor data and have a length of 500 time points each. 
The data is a recording of the sound of a motor engine running with abnormalities during the run-time.
\textsc{FordA} and \textsc{FordB} are utilized for binary classification tasks related to anomaly detection. 
The \textsc{FordA} dataset comprises 3601 samples in the training and 1320 in the test.
The \textsc{FordB} dataset comprises 3636 samples in the training and 810 in the test.
The \textsc{ElectricDevices} dataset has a length of 96 time points and seven classes.
The training set has a size of 8926, and the test set is 7711 time series long.

We explore a CNN and a ResNet architecture in our investigation. 
The CNN architecture is designed for the \textsc{FordA} dataset and the ResNet for the \textsc{ElectricDevices}. 
However, both architectures are applied to all three datasets.
The CNN architecture comprises three 1D convolutional layers and two fully connected layers. 
The Conv1D layers vary in kernel size and incorporate batch normalization and max pooling.
Rectified linear units (ReLU) are used as activation functions. 
The ResNet architecture follows the proposed architecture for time series classification by Fawaz et al.~\cite{ismailfawaz_deep_2019} with three 1D convolutional ResNet blocks and residual skip connections.
We use a batch size of 120 to train the model and employ the Adam optimizer~\cite{kingma_adam_2014}.
We further train both networks for 500 epochs without an earlier stopping and use the cross-entropy loss.

The performances of our models on the different datasets are as follows:
For the \textsc{FordA} dataset, our CNN (ResNet) achieves an accuracy of 0.99 (0.97) on the train and 0.89 (0.94) on the test set. 
For \textsc{FordB}, it is for the CNN (ResNet) 0.99 (0.98) on the train and 0.72 (0.80) on the test set.
For \textsc{ElectricDevies}, the CNN (ResNet) has 0.97 (0.90) and 0.71 (0.74) respectievly.
These results indicate a clear case of overfitting, as the models excessively adapt to the training data and struggle to generalize to unseen samples.
Despite our models' simplicity and overfitting, it is worth noting, as demonstrated by Ismail Fawaz et al.~\cite{ismailfawaz_deep_2019}, that these still exhibit performance comparable to state-of-the-art.

To illustrate the $ASI$ measure, we apply Saliency~\cite{simonyan_deep_2014}, DeepLIFT~\cite{shrikumar_learning_2017}, Integrated Gradients~\cite{sundararajan_axiomatic_2017}, GradientSHAP, and DeepLIFTShap~\cite{lundberg_unified_2017} on our selected datasets and our models and use the most suitable perturbation strategy for the datasets based on the results of Schlegel and Keim~\cite{schlegel_deep_2023} (OOD Low Sub for \textsc{FordA}, Zero Sub for \textsc{FordB}, and Global Max for \textsc{ElectricDevices}) using the implementation in Captum\footnote{\url{https://captum.ai/}}.
However, to mitigate some of the overfittings of Schlegel and Keim~\cite{schlegel_deep_2023}, we changed the architecture of the CNN by adding Batch Normalization~\cite{ioffe_batch_2015} and observed fewer flips of all training data.
Thus, the most suitable perturbation strategy could have changed, which needs further experiments.
We excluded Occlusion, ShapleyValueSampling, and KernelSHAP~\cite{lundberg_unified_2017} as these need further tuning and more runtime to process a whole dataset.
We plan to run an experiment with these in future works.
As we want to focus on attribution and time series, we use the weights $W = (0.5, 1, 0.5, 3)$ to visualize the projections and create a surface for the attributions.
However, we can also use the metric to compare the performance of the attributions on one sample..

\textbf{Hypothesis --}
For our experiment, we collected the following hypothesis:
The number of flips can be the same or similar for different attribution techniques~\cite{schlegel_deep_2023}, while $ASI$ can show a score more diverse than the flips or the quality metric change from the perturbation analysis~\cite{schlegel_towards_2019}.
$ASI$ can reveal highly relevant time points through a more diverse score and weighting.
Further, the $ASI$ score can reveal spurious correlations, e.g., also shown by Schlegel and Keim~\cite{schlegel_deep_2023} for a CNN on the \textsc{FordA} dataset when applied to the whole dataset.

\begin{figure}[h!tb]
    \centering
    \includegraphics[width=\textwidth,trim={0 124mm 18mm 2mm},clip]{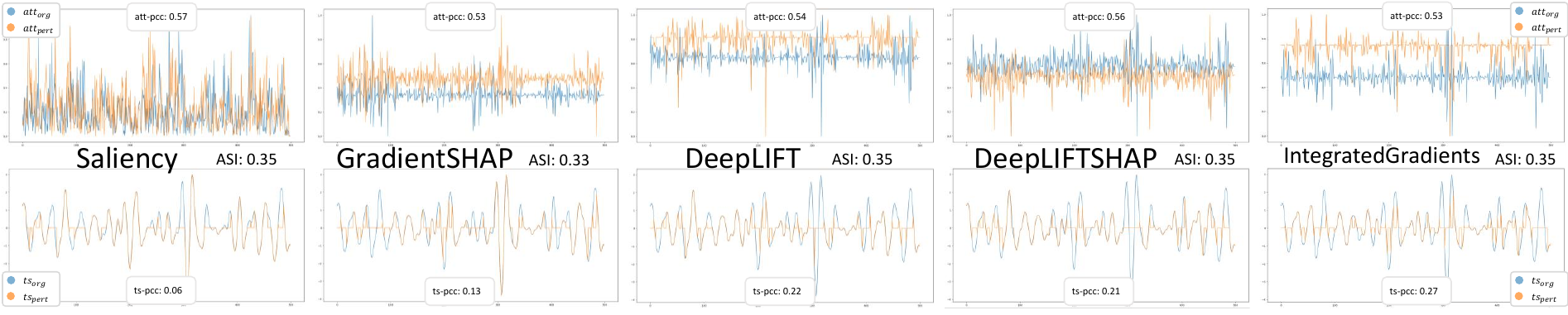}
    \caption{Different attribution techniques with the $ASI$ metric on a sample of the \textsc{FordA} test dataset for our CNN. GradientSHAP wins by a small margin against all others. The top row presents the attributions for the original time series and the perturbed instance time series. The bottom row shows the time series and the perturbed instances.}
    \label{fig:single_asi}
\end{figure}

\textbf{Results --}
\autoref{fig:single_asi} illustrates the results of $ASI$ for our selected attribution techniques and visualizes the time series and attributions.
The top row shows the attributions for the original time series instance in blue and those after the perturbation in orange.
Here, GradientSHAP and Integrated Gradients perform best on the attributions from the original to the perturbed instances by having a lower Pearson correlation coefficient.
However, inspecting the attributions based on a line plot can be tricky and misleading~\cite{schlegel_time_2021}.
The bottom row visualizes the time series instance in blue and the perturbed instance in orange.
Saliency does not change the time series much, while the others do pretty heavily.
The perturbation function here modifies relevant time points to zero and also, on both sides, two more points so that for every relevant variable, five time points are perturbed.
The score of~\autoref{eq:pccts} for Saliency is much lower because of the low change of the time series toward the perturbed instance.

\begin{figure}[h!tb]
    \centering
    \includegraphics[width=0.9\textwidth,trim={0 44mm 33mm 0},clip]{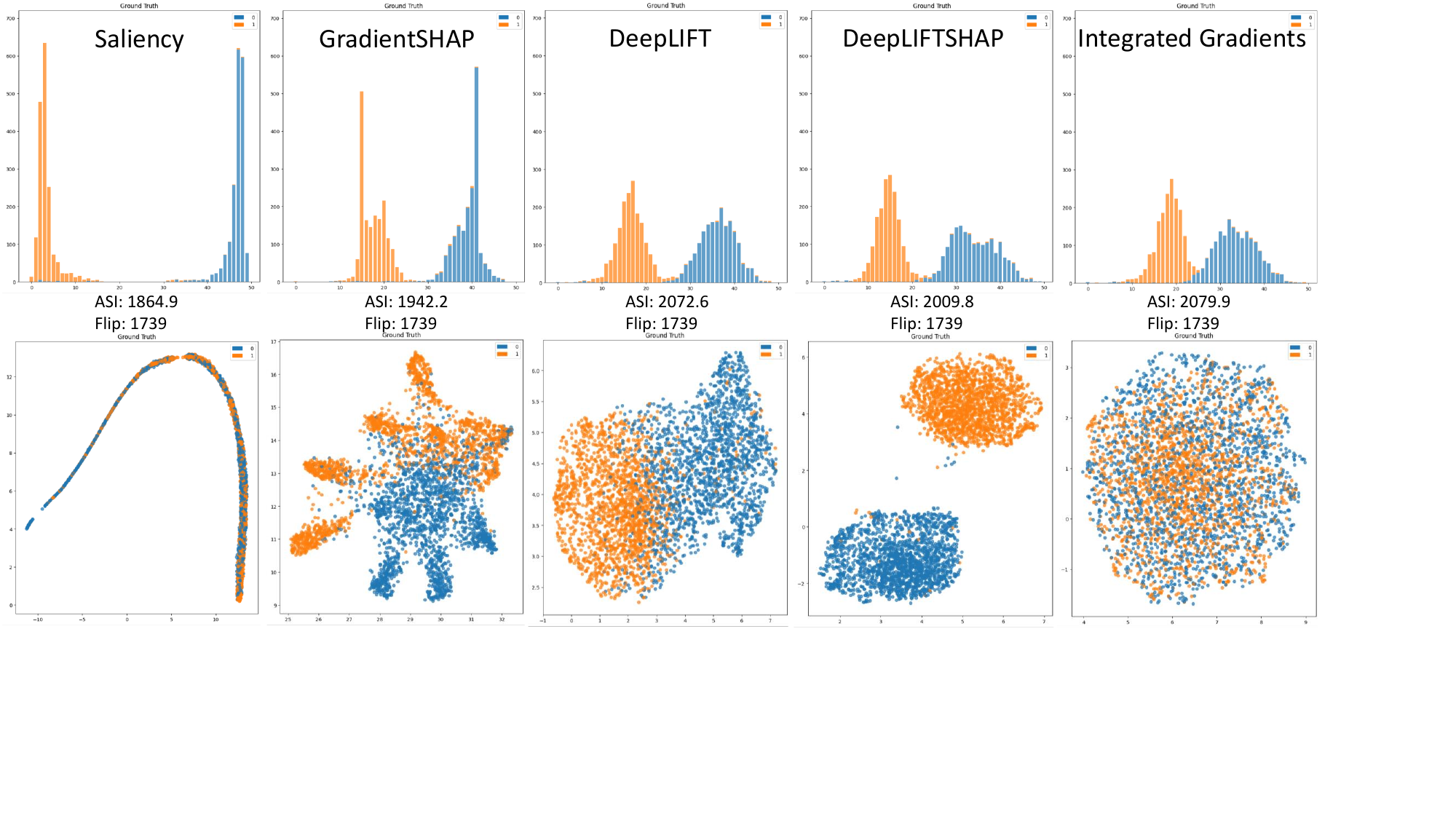}
    \caption{Top row: $ASI$ distribution scores for our CNN model on the \textsc{FordA} train dataset using Saliency, GradientSHAP, DeepLIFT, DeepLIFTSHAP, and Integrated Gradients. Bottom row: the projections of the attributions using UMAP with the class ground truth as coloring. The more Gaussian the $ASI$ distribution looks, and the farther the classes are separated, the better the projections are.}
    \label{fig:forda-train}
\end{figure}

\begin{figure}[h!tb]
    \centering
    \includegraphics[width=0.9\textwidth,trim={0 44mm 32mm 0},clip]{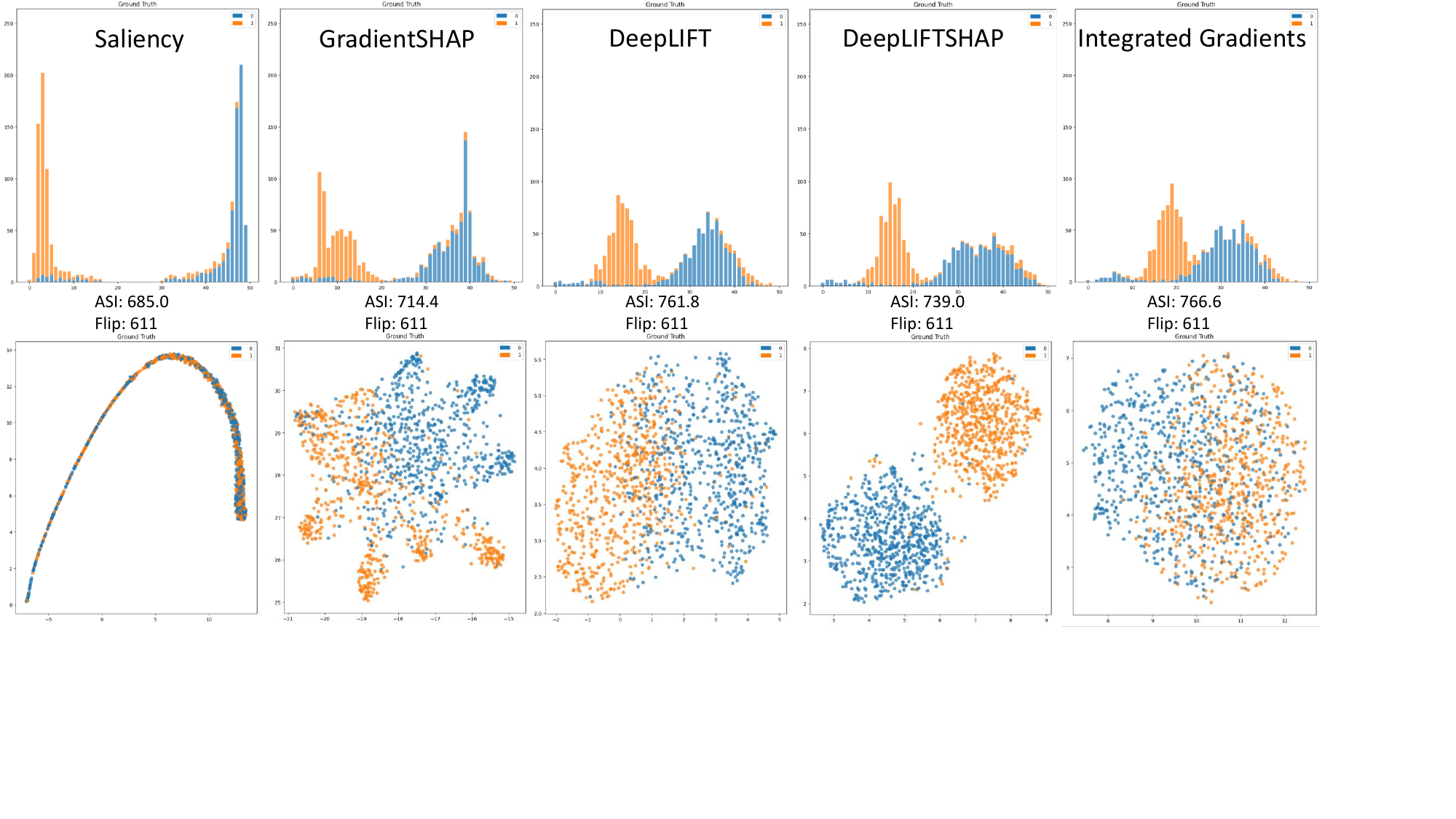}
    \caption{Top row: $ASI$ distribution scores for our CNN model on the \textsc{FordA} test dataset using Saliency, GradientSHAP, DeepLIFT, DeepLIFTSHAP, and Integrated Gradients. Bottom row: the projections of the attributions using UMAP with the class as coloring. }
    \label{fig:forda-test}
\end{figure}

To accept or reject our previously collected hypothesis, we need to focus not only on one sample but on whole datasets and their attributions.
Collecting all possible scores for $ASI$ on both the training and test datasets can be visualized in a distribution overview to gain insights into the attribution techniques (~\autoref{fig:forda-train} and~\autoref{fig:forda-test} top).
The $ASI$ score and the number of flips are also important (middle), and the projection of the attributions is necessary for our assumption (bottom).
At first glance, we can see that the $ASI$ score is more diverse than the number of flips in the figures, ~\autoref{fig:forda-train} and~\autoref{fig:forda-test} (static amount of flips and diverse $ASI$ score for the training data).
This result can be seen for all our datasets and confirms our first hypothesis.

We want to find interesting points in the attribution projection (~\autoref{fig:attribution-projection}).
Thus, we set for our weighting $W = (0.5, 1, 0.5, 3)$ to have a high distance between the attributions before and after the perturbation.
Based on such an assumption, a projection of the attributions demonstrates two separate clusters for a dataset with two classes.
Our focus works quite well for DeepLIFT and DeppLIFTSHAP on the \textsc{FordA} dataset, as seen in the training and test data (bottom in~\autoref{fig:forda-train} and~\autoref{fig:forda-test}).
For the training data (\autoref{fig:forda-train}), we can identify regions in the DeepLIFT projection where the attributions are mixed from both classes.
Samples in such regions are, in most cases, uncertain predictions with probabilities not entirely deciding on one class and containing interesting time points.
Analyzing these samples can lead to problems in the data and the model, e.g., wrong labels or overfitting.
We found that using the distribution of the $ASI$ score facilitates finding projections for such tasks.
The more Gaussian the $ASI$ distributions are, the better the projections.
If we further tune the weights, we can adjust the $ASI$ score and get an accessible number for the selection of the projection (e.g., increase attribution, decrease probability weights).

The ResNet has similar patterns for the \textsc{FordA} data and the distribution of the $ASI$ score.
The more Gaussian the $ASI$ distributions are, the better the projections.
However, for the ResNet, a more significant separation demonstrates worse projections.
Visualizing these projections also reveals different structures for the gradient-based techniques than for the CNN.
The projections are not split in clusters between the ground truth labels but are mixed heavily.
While the SHAP techniques are separated more into clusters.
Such a finding indicates that the skip-connections support the model to learn features more by heart than in the CNN, as attributions can be similar from one class to another. 
Slight changes can lead to changes in the prediction and the model fit; thus, similar gradients can lead to different predictions.
Adding SHAP helps to overcome the model fit and focuses on the prediction to have more robust attributions.

For both models, the FordA CNN and the ResNet, we see a flip only in one direction from a prediction of one to minus one.
Thus, the models only learn specific features necessary for predicting the one class and classify everything else as minus one.
Such a behavior can be seen as a Shortcut learning~\cite{geirhos_shortcut_2020}, which in some cases is undesirable if applied to real-world data and deployed in a critical environment.
Learning a more general pattern that can separate the classes and flipping not only one class can increase the robustness of the overall model.

The \textsc{FordB} data is a more complicated dataset for time series classification models as seen in the test score compared to, e.g., FordA~\cite{ismailfawaz_deep_2019}.
We found similar patterns to those of \textsc{FordA} for the CNN applied to \textsc{FordB}.
The more Gaussian the $ASI$ score is distributed, the more separated the ground truth clusters of the projections of the attributions are.
In the \textsc{FordB} case, this holds for the training and test data for nearly all attribution techniques except for Saliency, with Saliency having the lowest $ASI$ score.
For the ResNet, we also experience a similar pattern as for \textsc{FordA}.
SHAP techniques have larger $ASI$ scores but create separated clusters, while the gradient-based methods do not.

\textsc{ElectricDevices} captures seven classes and not only two, and thus is an even more challenging dataset.
The distribution of the $ASI$ score still separates into a double-peaked distribution.
However, the classes are not well divided into these peaked distributions anymore.
These build a more diverse pattern with some classes distributed into both peaks.
Here, our weighting seems quite misleading as the flips are less influential than the attributions; other methods are better than those identified by Schlegel and Keim~\cite{schlegel_deep_2023}.
If we focus on the best working technique identified by the lowest $ASI$ score, we get GradientSHAP.
Further, we also get similar-looking projections for the training and test data, which generally provides evidence that the method works well on the dataset.
However, inspecting the projections does not reveal our wanted intention of clusters separated by the classes.
Again, looking at the distributions and comparing them to Gaussians can help to select working projections.
The more Gaussian and equal-looking the double-peaks look, the better the projections.
If we change the weighting for this dataset to be more balanced, such as $W = (0.5, 1, 1, 2)$, we get an $ASI$ score corresponding to reasonable projections. 
In particular, DeepLIFTSHAP achieves the best $ASI$ score and creates clusters in the projection, capturing different classes in these clusters.

We further compared our results to the results of Schlegel and Keim~\cite{schlegel_deep_2023} and found some central differences.
We suggest investigating the Batch Normalization and the generation of attributions, as these drastically changed the projections.
We also suggest working with other projection techniques, such as PCA, and inspecting the visualizations, as UMAP can potentially demonstrate local minima and non-optimal solutions.
We also compared $ASI$ to the infidelity of Yeh et al.~\cite{yeh_in_2019} using the implementation of Captum on the CNN of the \textsc{FordA}.
The infidelity demonstrates more diverse scores between the train and the test data.
Thus, there is a change in the most suitable technique.
For the train data, these are Saliency and Integrated Gradients.
For the test data, these are DeepLIFTSHAP and Integrated Gradients.
We experienced Integrated Gradients as a less suitable technique with flawed projections by using $ASI$.
The complete results and further images can be found in the GitHub repository, which also can be used to reproduce the results and test other models and datasets.

\section{Conclusion \& Future Work}
We presented $ASI$, the attribution stability indicator, as a new measure for a perturbation analysis without using a quality metric on the whole dataset to compare attribution techniques.
$ASI$ incorporates five requirements we identified: class flip, prediction probability distribution change, high attribution distance, low time series distance, and user steering.
These requirements are incorporated using a binary function, the Jensen-Shannon distance, and the Pearson correlation coefficient.
As requirements can change from dataset to dataset, we include a weighting into $ASI$ to let users steer the factors.
We demonstrated $ASI$ on the \textsc{FordA}, \textsc{FordB}, and \textsc{ElectricDevices} datasets and discussed the influence of a weighting we selected for an $ASI$ analysis.
Based on the distributions over the score for whole datasets, we further presented how such an overview can lead to more information into the dataset and model using a projection.
While $ASI$ is quite general and can be easily extended to other data types, we tackle time series as the correlation between time series is widely used in literature~\cite{aghabozorgi_time_2015}, and thus the measure is easier to interpret and to understand.

However, we also identified shortcomings of $ASI$ and want to improve the measure with further additions.
One of these additions includes an OOD (out-of-distribution) check for the perturbed time series based on the attributions.
Through such a check, a plausibility check can be introduced into the perturbation analysis approach to include domain knowledge and a stronger focus on more real-looking time series.
Also, other distance functions can be added to incorporate more time series properties, such as the shape of the time series, as long as these are bounded between zero and one.
Another option is using a transformation function and the transformed data for a distance.
Further, a more extensive benchmark with, e.g., TimeReise~\cite{mercier_timereise_2022} can help to find fitting parameters for users.
Lastly, studying the attribution technique, perturbation strategy, and $ASI$ can help to analyze models in more depth.

\subsubsection*{Acknowledgements}
This work has been partially supported by the Federal Ministry of Education and Research (BMBF) in VIKING (13N16242).

%
%
%
\bibliographystyle{splncs04}
\bibliography{main}
\end{document}